\documentclass[11pt]{article}
\usepackage{nodalida2021}
\usepackage{times}
\usepackage{url}
\usepackage{latexsym}
\usepackage{graphicx}
\usepackage{arydshln}
\usepackage{color}
\usepackage{multirow}
\usepackage{verbatim}
\usepackage{wrapfig}
\usepackage{booktabs}
\usepackage{soul}
\usepackage{tikz}

\newcommand\mybox[2][]{\tikz[overlay]\node[fill=red!20,inner sep=1pt, anchor=text, rectangle, rounded corners=1mm,#1] {#2};\phantom{#2}}

\usepackage{multirow}
\aclfinalcopy 

\title{NLI Data Sanity Check: Assessing the Effect of \\ Data Corruption on Model Performance} 

\author{Aarne Talman\textsuperscript{*}\textsuperscript{$\dagger$}, Marianna Apidianaki\textsuperscript{*},  Stergios Chatzikyriakidis\textsuperscript{$\ddagger$}, J\"org Tiedemann\textsuperscript{*}\\~\\
\textsuperscript{*}Department of Digital Humanities, University of Helsinki\\
\texttt{\{name.surname\}@helsinki.fi}\\
\textsuperscript{$\dagger$}Basement AI\\
\textsuperscript{$\ddagger$}CLASP, Department of Philosophy, Linguistics and Theory of Science, 
University of Gothenburg\\
\texttt{\{name.surname\}@gu.se}}

\date{}

\begin{document}
\maketitle

\begin{abstract}
Pre-trained neural language models give high performance on natural language inference (NLI) tasks. But whether they actually understand the meaning of the processed sequences  
remains unclear. We propose a new diagnostics test suite which allows to assess whether a dataset constitutes a good testbed for evaluating the  models' meaning understanding capabilities. We specifically apply controlled corruption transformations to  widely used  benchmarks (MNLI and ANLI), which  involve removing entire word classes and often lead to non-sensical sentence pairs.  If model accuracy on the corrupted data remains high, then the dataset is likely to contain statistical biases and artefacts that guide prediction. Inversely, a large decrease in model accuracy indicates that the original dataset provides a proper challenge to the models' reasoning capabilities. Hence, our proposed controls can serve as a crash test for developing high quality data for NLI tasks.
\end{abstract}

\section{Introduction}

Assessing the natural language inference (NLI) and 
understanding (NLU) capabilities of  a model poses numerous challenges, one of which is 
 the quality and composition of the data 
used for evaluation. Popular NLI datasets \citep{snli,marelli-etal-2014-sick} contain annotation artefacts and statistical irregularities  that can be easily grasped by a model during training and guide prediction, even if the model has not acquired the   
knowledge needed to perform  this kind of reasoning. 
Notably, recent work shows that major modifications such as word shuffling do not hurt BERT's \cite{bert} NLU capabilities much, mainly due to individual words' impact on prediction 
\cite{Phametal:2020}. To the contrary,  
small tweaks or perturbations in the data, such as replacing words with mutually exclusive co-hyponyms and antonyms \cite{breakingNLI} or changing the order of the two sentences \cite{wang2019}, has been shown to hurt the performance of NLI models.   

\begin{table}[t!]
\small
\centering
        \begin{tabular}{p{.3cm} p{3.2cm}  p{2.5cm}} 

& \parbox{3.2cm}{\centering \bf Premise} & \parbox{2.5cm}{\centering \bf  Hypothesis} \\ 
\hline
\multirow{1}{*}{\rotatebox[origin=c]{90}{Contradiction}} & \sl \vspace{.5mm} He was hardly more than five \mybox[fill=red!20]{\st{feet}}, four \mybox[fill=red!20]{\st{inches}}, but carried himself with great \mybox[fill=red!20]{\st{dignity}}. \vspace{1.5mm} & \sl \vspace{.5mm}  The \mybox[fill=red!20]{\st{man}} was 6 \mybox[fill=red!20]{\st{foot}} \mybox[fill=red!20]{\st{tall}}.\\ 
\hdashline
\multirow{1}{*}{\rotatebox[origin=c]{90}{\parbox{2cm}{\centering Entailment}}} & \sl \vspace{.5mm}  Two \mybox[fill=red!20]{\st{plants}} died on the long \mybox[fill=red!20]{\st{journey}} and the third one found its way to \mybox[fill=red!20]{\st{Jamaica}} exactly how is still shrouded in \mybox[fill=red!20]{\st{mystery}}. &\sl \vspace{.5mm}  The third \mybox[fill=red!20]{\st{plant}} was a different \mybox[fill=red!20]{\st{type}} from the first two.\\ 
\hdashline
\multirow{1}{*}{\rotatebox[origin=c]{90}{\parbox{2cm}{\centering Neutral}}} &  \sl \vspace{.5mm}  In a \mybox[fill=red!20]{\st{couple}} of \mybox[fill=red!20]{\st{days}} the \mybox[fill=red!20]{\st{wagon}} \mybox[fill=red!20]{\st{train}} would head on north to \mybox[fill=red!20]{\st{Tucson}}, but now the \mybox[fill=red!20]{\st{activity}} in the \mybox[fill=red!20]{\st{plaza}} was a \mybox[fill=red!20]{\st{mixture}} of \mybox[fill=red!20]{\st{market}} \mybox[fill=red!20]{\st{day}} and \mybox[fill=red!20]{\st{fiesta}}. &\sl \vspace{.5mm}  They were \mybox[fill=red!20]{\st{south}} of \mybox[fill=red!20]{\st{Tucson}}. \\ 
            \hline
        \end{tabular}
    \caption{\label{table:example_train_data} 
    Sentence pairs from a corrupted 
    MNLI training dataset 
    where nouns have been removed. 
    } 
\end{table} 

Motivated by this situation, our goal is to contribute a new suite of diagnostic tests that can be used to assess the quality of an NLU benchmark.  In particular, we conduct a series of controlled experiments where a set of data corruption transformations are 
applied to the widely used MNLI \citep{multinli} and ANLI  \citep{nieetal2020} datasets, and explore their impact on fine-tuned BERT and ROBERTa \citep{liu2019roberta} model performance. 
The obtained results provide evidence that can reveal the  
quality of a dataset: 
 Given that the transformations 
seriously affect the quality of NLI sentences, going as far as making them unintelligible (cf. examples in Table \ref{table:example_train_data}), a decrease in  performance for models fine-tuned on the corrupted dataset would be expected. High performance would, instead, indicate the presence of biases and other artefacts in the dataset which 
guide models' predictions. 
This situation would be indicative of a low quality dataset, i.e. one we cannot rely upon 
to draw safe conclusions about a model's NLI capabilities. 

Bringing in additional evidence to the debate on problematic NLI evaluation setups and how poorly they represent the real inference capabilities of the tested models, our proposed  diagnostics allow  to evaluate the quality of datasets 
by assessing how artefact and bias-free they are, and hence the extent to which they can be trusted for evaluating NLI models' language reasoning capabilities. We consider this step highly important for estimating the quality of existing benchmarks and interpreting model 
results accordingly, and for guiding the development of new datasets addressing inference and reasoning. We make our code and data available in order to promote the adoption of these diagnostic tests and facilitate their application to new datasets.\footnote{\url{https://github.com/Helsinki-NLP/nli-data-sanity-check}}

\section{Related Work}\label{section:related_work}

A well-known problem of  NLU  evaluation benchmarks is that the proposed tasks are often  solvable by simple heuristics 
\citep{hewitt-liang-2019-designing}. This is mainly due to the presence of linguistic biases in the datasets, which  make prediction easy 
\citep{lai-hockenmaier-2014-illinois,poliak2018}. 
Notably, 90\% of the hypotheses that denote a contradiction in the original SNLI dataset \citep{snli}  contain the verb {\it sleep} and its variants ({\it sleeping}, {\it asleep}) which serve to  mark a contrast with an  activity described in the premise (e.g., {\it My sister is playing} $\rightarrow$ {\it My sister is sleeping}); while contradictions in SICK \citep{marelli-etal-2014-sick}  are often marked by explicit negation. This latter issue also exists in SNLI and MNLI as spotted by  \newcite{gururangan-etal-2018-annotation}, where negation is highly indicative of contradiction,  
and generic nouns (e.g., {\it animal}, {\it something}) of entailment. These grammatical or lexical cues  are easily grasped by the models during training and help them correctly predict  the relationship between two sentences, but this does not  
mean that the models are capable of performing 
this type of  reasoning. 
Notably, due to these annotation artefacts and statistical irregularities, it is possible even for hypothesis-only NLI models (i.e. models that are fine-tuned only on the hypotheses without  access to the premises) to make correct predictions \cite{poliak2018}. 

Recent work shows that state-of-the-art NLU models are not very sensitive to word order which, however, is one of the most important characteristics of a sequence \cite{Phametal:2020}. Specifically, performance of BERT-based classifiers fine-tuned on GLUE tasks \cite{wang-etal-2018-glue} remains relatively high 
after randomly shuffling input words. This is mainly explained by the contribution of each individual word 
which remains unchanged after its context is shuffled. 
Superficial cues such as the sentiment of keywords in sentiment analysis, or the word level  
similarity between sentence pairs in NLI, allow BERT-based models to make correct decisions even when tokens are arranged in random orders, suggesting that many GLUE tasks are not really challenging them to understand the meaning of a sentence. 

To the contrary, when simple heuristics do not suffice to solve the NLI task, NLI systems seem to be more prone to breaking. This is for example what happens when swapping the test and training datasets of different benchmarks (i.e. training on one NLI dataset and testing on an other)  \cite{talman-chatzikyriakidis-2019-testing}.   \citet{wang2019} report problems in performance when  
the premise and the hypothesis are swapped. The idea is that 
the label of contradicting or neutral pairs should remain the same in the case of a swap, in contrast to 
 entailment pairs where a different label should be proposed after the swap. This would be expected because  entailment is a directional relationship, while contradiction is symmetric.\footnote{More explicitly, for contradiction, the idea is that 
when $A \to \neg B$ (i.e.  B contradicts A), then, by contraposition, 
$B \to \neg A$ also holds (A contradicts B).} \citet{wang2019} test 
various models with respect to this diagnostic and observe a a significant drop in performance  (i.e. predicted labels change) when the contradicting and neutral pairs are swapped. 
The models' behaviour seems more reasonable when these are tested on the swapped entailment pairs, where all but one models correctly predict a different label. 
In the light of these results, the authors propose the swapping method 
as a sanity check for NLI models. 

The low quality of existing datasets and the impressively high performance of NLI systems, as measured on these benchmarks, have  sparked a new research direction where the goal is to propose  
new more challenging and artefact-free datasets. 
The ANLI  dataset, for example, 
was built precisely with the goal to eliminate annotation artefacts \citep{nieetal2020}.
The authors claim that this dataset  
is much less prone to annotation artefacts compared to previous benchmarks, 
as suggested by the lower prediction accuracy for models  
fine-tuned on the ANLI hypothesis-only dataset. Although there still seems to be space for improvement 
(accuracy is around 0.5, i.e. 
well above chance), 
the reported findings are promising. Specifically, the performance is lower than on the 
hypothesis-only SNLI/MNLI datasets, showing that the dataset contains less artefacts that can guide prediction. 
ANLI is thus a natural candidate to further test our hypotheses, as it claims to remedy for a number of the  shortcomings of earlier NLI datasets. 

Lessons learnt from previous work on designing reliable linguistic probing tasks \citep{hewitt-liang-2019-designing} and the overfitting problems of NLI models discussed above,  demonstrate the importance of systematic sanity checks  
like the ones we propose in this paper. Our dedicated control tasks specifically allow to determine whether a dataset triggers the models' reasoning capabilities or, instead, allows them to rely on statistical biases and annotation artefacts for prediction. We use the quality of the predictions made by models fine-tuned and tested on corrupted data as a proxy to evaluate data quality.

\section{Datasets}
\label{datasets}
\subsection{The Multi-Genre NLI (MNLI) Corpus}

We carry out our experiments on the  Multi-Genre Natural Language Inference (MNLI) corpus \citep{multinli}.  
MNLI contains 433k human-written sentence pairs labeled as ``entailment'', ``contradiction'' and ``neutral''. The corpus includes sentence pairs from ten distinct genres of written and spoken English,\footnote{MNLI text genres:  Two-sided in person and telephone conversations ({\sc Face-to-face, Telephone}); content from public domain government websites ({\sc Government});  letters from the Indiana Center for Intercultural Communication of Philanthropic Fundraising Discourse ({\sc Letters}); the public report from the National Commission on Terrorist Attacks Upon the United States ({\sc 9/11}); non-fiction works on the textile industry and child development (OUP); popular culture articles ({\sc Slate}); travel guides ({\sc Travel}); short posts about linguistics for non-specialists ({\sc Verbatim}); {\sc Fiction}.} making it possible to 
approximate a wide variety of ways in which modern standard American English is used, and supplying a setting for evaluating cross-genre domain adaptation. 
All ten genres appear in the test and development sets, but only five are included in the training set. The MNLI development and test sets have been divided into ``matched'' and ``mismatched'': The former includes only sentences from the same genres found in 
the training data, and the latter includes sentences from the remaining genres not present in the training data. For our experiments, we use the development sets as our evaluation data since the annotated test sets are not publicly available.

\subsection{The Adversarial NLI (ANLI) Corpus}

The Adversarial NLI benchmark (ANLI) \citep{nieetal2020} was specifically designed to address some of the shortcomings of the previous NLI datasets. 
ANLI contains three 
datasets (rounds), R1, R2 and R3. Each dataset was collected using a human-and-model-in-the-loop approach, and they progressively increase 
in difficulty and complexity. The annotators were shown a 
context (premise) and a target label, and were asked to 
propose a hypothesis that would lead a model to miss-classify the label. For R1, the 
model that the annotators were asked to deceive was BERT-Large, while for R2 and R3, it was RoBERTa. For R3, the contexts were selected from a wider set of sources.\footnote{The contexts for R1 and R2 consist of sentences retrieved from Wikipedia. In R3 the contexts are retrieved from Wikipedia, News (Common Crawl), fiction, The Children's Book Test (CBT), formal spoken text and procedural text extracted from WikiHow.} 
The corpus also includes label explanations provided by the annotators. Each round (R1-R3) contains training, development and test data. 

ANLI is a relatively small dataset. R1 consists of only 16,946 training examples, 
1,000 development and 1,000 test examples. 
R2 is slightly larger, it contains 
45,460 training examples and the same number of development and test examples as R1. 
Finally, R3 
contains 100,459 training examples and slightly larger development and test sets (1,200 each).

\subsection{Systematic NLI Data Corruption}\label{section:datasets}

We create modified versions of the MNLI training and evaluation data by applying a set of controlled transformations to the original  dataset. We call these two sets MNLI {\sc  Corrupt-Train} and {\sc Corrupt-Test}, respectively. We specifically remove words of specific word classes  after tagging the texts with universal part of speech (POS) tags using the NLTK library and the averaged perceptron tagger.\footnote{\url{https://www.nltk.org/}.}  
In the obtained {\sc MNLI-noun} training dataset, for example, all nouns in the original MNLI training data have been removed. We furthermore create training data following the inverse process, i.e. 
keeping only words of specific  classes and removing the others. For example, the {\sc noun+verb} dataset contains only nouns and verbs from the original MNLI sentences. 

We similarly create 
the {\sc Corrupt-Test} set by removing words of specific word classes 
from the MNLI-matched development dataset, or keeping these and removing the rest. 
Table \ref{table:datasets} in the Appendix contains statistics about the training and evaluation datasets obtained 
after applying each transformation. 
Finally, we combine the original MNLI and the corrupted training datasets together. 
{\sc MNLI-AllDrop} 
contains the following training sets: 
 MNLI (original), {\sc-num}, {\sc-conj}, {\sc-adv},  {\sc-pron}, {\sc-adj}, {\sc-det}, {\sc-verb}, {\sc-noun}. 
 
 We use  ANLI as an example of a high quality dataset, and create  {\sc  ANLI-Corrupt-Test} 
by applying all the -{\sc pos} transformations on the 
ANLI test sets. Table \ref{table:anli-stats} in the Appendix contains statistics about the different {\sc  ANLI-Corrupt-Test} datasets. 
To test the effect of corrupting the training data used in ANLI experiments \citep{nieetal2020}, we also create a training set that consists of the SNLI, MNLI, FEVER and ANLI training data with all the occurrences of nouns removed ({\sc  ANLI-Corrupt-Train}). 

We test the performance of BERT on the corrupted MNLI data, and that of RoBERTa on the corrupted ANLI data, and compare the results to those obtained using the original datasets. 
 We expect models fine-tuned on corrupted data -- where important information is missing and sentences often do not make sense -- to perform poorly compared to the same models fine-tuned on the original data. High performance of models fine-tuned on these highly problematic data would indicate that the models leverage clues (biases and artefacts) that are present in the data, instead of performing reasoning operations. Inversely, low model performance  would suggest that they are unable to reason using these corrupted data, and that the data do not contain artefacts that would guide prediction in this setting. 

\section{Models}

\begin{table*}[ht!]
\small
\centering
\scalebox{0.95}{
        \begin{tabular}{l c c | c c | c c}

\bf Data &  \bf {\sc Corrupt-Train} 
&\bf $\Delta$ & \bf {\sc Corrupt-Test}  
&\bf $\Delta$ & \bf {\sc Corrupt-Train and Test} 
&\bf $\Delta$\\
            \hline
 {\sc mnli-num} & 82.37\% & -1.37 & 81.71\% & -2.03 & 81.87\% & -1.87\\
{\sc mnli-conj} & 83.09\% & -0.65 & 82.75\% & -0.99 & 83.10\% & -0.64\\
{\sc mnli-adv} & 80.21\% & -3.53 & 72.41\% & -11.33 & 75.69\% & -8.05\\
{\sc mnli-pron} & 83.27\% & -0.47 & 81.98\% & -1.75 & 82.65\% & -1.09\\
{\sc mnli-adj} & 81.67\% & -2.07 & 74.61\% & -9.13 & 76.44\% & -7.30\\
{\sc mnli-det} & 83.15\% & -0.59 & 79.29\% & -4.44 & 81.32\% & -2.42\\
{\sc mnli-verb} & 81.40\% & -2.34 & 73.96\% & -9.78 & 76.30\% & -7.44\\
{\sc mnli-noun} & 80.72\% & -3.02 & 69.80\% & -13.94 & 73.38\% & -10.35\\
{\sc mnli-noun-pron} & 79.74\% & -4.00 & 68.41\% & -15.33 & 72.14\% & -11.60\\
\hdashline
{\sc noun+pron+verb} & 72.55\% & -11.19 & 54.59\% & -29.15 & 62.18\% & -21.56\\
{\sc noun+adv+verb} & 67.58\% & -16.16 & 62.58\% & -21.16 & 67.58\% & -16.16\\
{\sc noun+verb} & 71.14\% & -12.60 & 52.90\% & -30.84 & 61.31\% & -22.43\\
{\sc noun+verb+adj} & 75.54\% & -8.20 & 61.90\% & -21.84 & 68.20\% & -15.54\\
{\sc noun+verb+adv+adj} & 79.81\% & -3.93 & 71.81\% & -11.93 & 76.29\% & -7.45\\
            \hline
        \end{tabular}
        }
    \caption{\label{table:results} Prediction accuracy (\%) for the BERT-{\tt base} model 
    fine-tuned on 
    {\sc Corrupt-Train} 
    and tested on the original MNLI-matched evaluation (dev) set (columns 2 and 3);  fine-tuned on the original MNLI data and tested on {\sc Corrupt-Test}; 
    fine-tuned on {\sc Corrupt-Train}  and tested on {\sc Corrupt-Test} 
    (columns 6 and 7). 
    The delta shows the difference in accuracy compared to the model fine-tuned on the original MNLI training set and evaluated on the MNLI-matched development set (83.74\%).
    }
\end{table*}

We use Google's original TensorFlow implementation\footnote{\url{https://github.com/google-research/bert}} of the uncased 768-dimensional BERT model (BERT-{\tt base}), a transformer model that learns representations via a bidirectional encoder \citep{bert}. 
BERT was pre-trained using a Masked Language Model (MLM or cloze) task where some percentage of the input tokens are masked at random, and the model needs to predict these masked tokens; and on a Next Sentence Prediction (NSP) task, where it receives pairs of sentences(A, B) as input and learns to predict if B follows A 
in the original document. Sentence B in (A, B) 
is 50\% of the time the actual sentence that follows A, and 50\% of the time it is a random sentence from the training corpus. NSP increases the model's ability to capture the relationship between two sentences, which is the core task in NLI and Question Answering.

Variants of the BERT model achieve very high performance on NLU tasks, surpassing the human baseline on GLUE \cite{wang-etal-2018-glue} and reaching near-human performance on the challenging SuperGLUE dataset \cite{NEURIPS2019_4496bf24}.  
For each experiment, we fine-tune BERT for ten  epochs on the original MNLI training dataset or its transformed versions described in Section \ref{datasets}, using a batch size of 100  (unless explicitly stated).

For the experiments on the ANLI benchmark, we apply the RoBERTa-{\tt large} model, a variant of BERT 
which has much higher performance than BERT on the GLUE and SuperGLUE benchmarks.\footnote{The  modifications in RoBERTa include training the model longer, with bigger batches, over more data and on longer sequences. The pre-training approaches has also been modified compared to BERT: 
The next sentence prediction objective is removed and 
dynamic masking is introduced. This results in different tokens being masked across 
training epochs.} We use the training and evaluation scripts provided by 
\citet{nieetal2020}. 
\footnote{\url{https://github.com/facebookresearch/anli}} We fine-tune the model for two epochs using a batch size of 16. 

\section{Evaluation}

\subsection{ {\sc Corrupt-Train} and Original Test}
\label{corrupt-train}

We 
evaluate the performance of the BERT model when fine-tuned on each of the 14 training sets in MNLI {\sc Corrupt-Train}.  
We measure the models' prediction accuracy on  the original MNLI-matched development dataset, which serves as our 
test set. The results given in the first column of Table \ref{table:results} 
show that removing all the occurrences of a specific 
word class from the MNLI training data has a surprisingly low impact on BERT's performance, which remains high.  
As expected, the biggest decrease is observed when content words are removed, with adverbs having the largest impact (-3.53), followed by nouns (-3.02) and verbs (-2.34). 
Interestingly, the number of nouns is 4.5 times higher than the number of adverbs in the dataset,  suggesting that the latter have a larger impact on NLI prediction. 
The 
small drop in accuracy observed across the board is, however, 
highly surprising. Arguably, sentences with nouns removed make very little sense to humans (cf. Table \ref{table:example_train_data}).\footnote{Cf. Table \ref{table:example_correct} in the Appendix for examples of corrupted sentence pairs from the {\sc mnli-noun} test set for which BERT has made a correct prediction.} 
The observed high performance of BERT on these problematic data might be due to the 
knowledge  about gap filling and Next Sentence Prediction acquired by the model during pre-training,  which it can still leverage and combine with other cues  in the training and test data for  prediction.

\begin{table*}[ht!]
\centering
\scalebox{0.9}{
        \begin{tabular}{l c c  }
\bf Training Data &\bf MNLI-matched (dev) &\bf MNLI-mismatched (dev) \\
            \hline
MNLI & 83.74\% & 83.76\% \\
MNLI-{\sc AllDrop} &\bf 84.09\% &\bf 84.30\% \\
            \hline
        \end{tabular}}
    \caption{\label{table:train_big} Comparison of prediction accuracy (\%) for BERT-{\tt base} models fine-tuned on the original MNLI training set and 
    on {\sc MNLI-AllDrop}, and tested on 
    the original MNLI evaluation (dev) sets.}
\end{table*}

\subsection{Evaluation on {\sc Corrupt-Test}} 

\paragraph{Models fine-tuned on original data.} We evaluate the performance of the BERT model fine-tuned on the original MNLI training data, on our {\sc Corrupt-Test} data. 
The middle columns of Table \ref{table:results} show the experimental results on the different 
{\sc Corrupt-Test} datasets, and the difference (delta) 
from the results on the original (unmodified) MNLI-matched development set.

\begin{table*}[!ht]
\small
\centering
        \begin{tabular}{l c c | c c | c c}
\bf Data & \bf {\sc  Corrupt-Test} R1 & \bf $\Delta$  & \bf{\sc  Corrupt-Test} R2 &\bf $\Delta$ & \bf{\sc  Corrupt-Test} R3 &\bf $\Delta$ \\
            \hline
{\sc anli-conj } & 70.2\% & -3.6 & 49.0\% & 0.1 & 46.5\% & 2.1 \\
{\sc anli-pron} & 69.6\% & -4.2 & 49.7\% & 0.8 & 45.0\% & 0.6 \\
{\sc anli-det} & 69.5\% & -4.3 & 49.4\% & 0.5 & 45.0\% & 0.6 \\
{\sc anli-adv} & 67.1\% & -6.7 & 49.6\% & 0.7 & 43.8\% & -0.6 \\
{\sc anli-adj }& 60.2\% & -13.6 & 45.1\% & -3.8 & 45.0\% & 0.6 \\
{\sc anli-num }& 58.7\% & -15.1 & 43.8\% & -5.1 & 45.1\% & 0.7 \\
{\sc anli-verb }& 54.6\% & -19.2 & 44.7\% & -4.2 & 39.3\% & -5.1 \\
{\sc anli-noun} & 43.7\% & -30.1 & 36.0\% & -12.9 & 32.4\% & -12.0 \\
            \hline
        \end{tabular}
    \caption{\label{table:anli-test_results} Prediction accuracy (\%) for the  RoBERTa-{\tt large} model 
    on the {\sc Corrupt} R1, R2 and R3 test sets. 
    Delta shows the  difference in accuracy compared to the state-of-the-art results reported by \citet{nieetal2020} on the original test sets, R1: 73.8\%, R2: 48.9\% and R3: 44.4\%.} 
\end{table*}

We observe a similar pattern as in the previous experiment. Removing content words (nouns, verbs and adverbs) has the strongest impact on model accuracy, 
whereas eliminating conjunctions and numerals has only a small impact on the results. The decrease in prediction accuracy 
observed in this setting is more important than in the evaluation of models fine-tuned on {\sc Corrupt-Train} and tested on unmodified data. 
Nevertheless, the fact that BERT can still predict the correct label with fairly high accuracy in cases where all the nouns 
or verbs are removed is 
surprising, since these transformations often lead to almost unintelligible sentence pairs (cf. examples in Table \ref{table:example_train_data} 
in the paper and Table \ref{table:example_correct} in the Appendix). 
Since inference in such non-sensical sentences cannot rely on meaning, our explanation for the models' performance is that they leverage other clues and biases that remain in the sentences after corruption 
for prediction. 
Note that the models tested in this setting were fine-tuned on the original MNLI data.  
We believe that during this stage 
the model acquires 
knowledge about possible sequence pairs, including the artefacts and other clues therein.

\paragraph{Models fine-tuned on {\sc Corrupt-Train}.}  
We evaluate the performance of BERT models fine-tuned on {\sc Corrupt-Train}, 
on {\sc Corrupt-Test}. The results of these experiments are shown in the last two columns of Table \ref{table:results}. We observe again a similar pattern in terms of relative importance of the different word classes, with  
content words having the biggest impact. What is definitely surprising in this case is that the drop in performance 
is smaller than the one observed for the models trained on the original data and tested on {\sc corrupt-test}, suggesting that the model relies on data artefacts even more in this setting.

\subsection{{\sc MNLI-AllDrop} Evaluation}

Motivated by the small decrease 
in prediction accuracy observed 
when removing specific word classes from the training data (cf. Section \ref{corrupt-train}), we also fine-tune the  model on a large dataset combining the different {\sc Corrupt-Train} sets and the original MNLI training set. 
The BERT fine-tuning code is shuffling the provided examples, so our goal here is to explore whether seeing sentence pairs where  words of different classes are missing (e.g., sentences without verbs following 
sentences that contain no nouns) confuses the model.

The results of this experiment are shown in  Table \ref{table:train_big}. They indicate that 
removing occurrences of different word classes from the sentences during training can act as a regularisation technique and, hence improve the model performance. We observe a small increase (+0.35) when evaluated on the original MNLI-matched development data, and an increase of 0.56  when evaluated on the original MNLI-mismatched development data.

\subsection{Evaluating on ANLI}

In order to demonstrate 
that systematic data corruption can be a useful 
diagnostic for evaluating benchmark quality, we conduct additional experiments on the 
ANLI test set  \citep{nieetal2020}. 
The results for 
the RoBERTa-{\tt large} model fine-tuned on the original   
training sets and 
evaluated on {\sc Corrupt-Test} R1, R2 and R3 data 
are given in Table \ref{table:anli-test_results}. 

As expected, we observe a clear drop in accuracy for the datasets where content-bearing words  are removed ({\sc -nouns, -verbs}), and a relatively small drop when function words  are missing ({\sc -conj, -det}), but only in R1. 
However, the fact that accuracy on the R2 and R3 datasets improves after some corruption transformations are applied 
({\sc anli-pron, -conj, -det}) is an interesting finding. 
A possible explanation is that as the sentences (especially the premises) are much longer in ANLI compared to other NLI datasets, removing non-content-bearing words makes it easier for the model to grasp the essential
information for making correct predictions. The large drop in accuracy when 
nouns and verbs are removed supports our hypothesis regarding 
the superior quality of the ANLI corpus compared to MNLI, suggesting that the dataset contains less artefacts on which the model can base prediction after corruption. 

We also compare the 
results reported by \citet{nieetal2020} for the RoBERTa-{\tt large} model 
to the ones obtained with the model  
 fine-tuned on 
the {\sc anli-noun}  training set.\footnote{This corresponds to  {\sc mnli+snli+fever+anli} with all nouns  removed.}  We measure the model's prediction accuracy on the original R1, R2 and R3 test sets, and report the results in Table \ref{table:anli_results}. The  drop in prediction accuracy is significantly larger than that observed on the MNLI data.  
Hence, the data corruption procedure reveals the improved quality of the ANLI data set as a benchmark for NLU. However, the fact that the model is able to predict the correct label with 57.6\% accuracy for ANLI R1 highlights that even with this dataset the model learns some factors from the data that it is able to use when predicting the label for a pair, even when 
the training sentences do 
not make much sense. These results further demonstrate the importance of carefully running diagnostics such as ours to assess the use of a new benchmark in NLU tasks.

\begin{table}[t!]
\small
\centering
        \begin{tabular}{l l c c c}
\bf Training data & \bf R1 &\bf R2 & \bf R3\\
            \hline
 {\sc anli} 
 &   73.8\% & 48.9\% & 44.4\% \\
 {\sc anli-noun} &   57.6\% & 40.3\%  &  41.0\% \\
            \hline
        \end{tabular}
    \caption{\label{table:anli_results} Prediction accuracy (\%) for RoBERTa-{\tt large} on the {\sc anli-noun} dataset. 
    Comparison to the 
    results of \citet{nieetal2020} on the original ANLI dataset. ANLI contains {\sc mnli, snli, fever} and {\sc anli}.} 
\end{table}

\section{Discussion}

The question of whether  current state-of-the-art  neural network models  that beat human performance in  NLU tasks actually understand language is currently much debated.
Our proposed corruption transformations often lead to sentences that make very little sense.
Nevertheless, 
we observe that BERT performs surprisingly well in these experiments. This indicates that rather than understanding the meaning of the sentences and the semantic relationship between them, the models are able to pick up on other cues from the data that allow them to make correct predictions.

Our proposed diagnostics tests are  useful devices for assessing the quality of a dataset as a testbed for evaluating  models' language understanding capabilities. In our experiments, they  demonstrate the superior quality of a NLI dataset (ANLI) over another (MNLI). We test this finding in an additional experiment where we apply the word shuffling mechanism of \citet{Phametal:2020} on the ANLI data, which 
was shown to not deteriorate BERT-based model performance on the GLUE tasks. 
Our results in Table  \ref{table:anli-suffle_results} show  that this procedure significantly hurts model accuracy on  ANLI, and bring in additional evidence supporting the superior quality of this dataset over MNLI (which is part of 
the GLUE benchmark). 

\begin{table}[t!]
\small
\centering
        \begin{tabular}{l c c c}
\bf Test set &\bf R1 &\bf R2 &\bf R3\\
\hline
{\sc anli} & 73.8\% & 48.9\% & 44.4\%\\
{\sc anli-shuffle}-n1 & 35.5\% & 33.8\% & 36.0\%\\
{\sc anli-shuffle}-n2 & 45.4\% & 39.8\% & 37.1\%\\
{\sc anli-shuffle}-n3 & 49.4\% & 40.7\% & 38.4\%\\
\hline
        \end{tabular}
    \caption{\label{table:anli-suffle_results} Prediction accuracy (\%) for   RoBERTa-{\tt large}  
    after word shuffling 
    \citep{Phametal:2020}. Comparison to 
    results obtained on the original ANLI dataset \cite{nieetal2020}. 
    The {\sc anli-shuffle}-n1/n2/n3 test sets 
    contain shuffled n-grams, with  $n=\{1,2,3\}$ respectively.}
\end{table}

Our test suite can be seen as an additional ``crash test'' for assessing the quality of benchmark datasets that address common-sense reasoning. It falls in the same line as 
work that highlighted problems of earlier datasets and resulted in the creation of ANLI. 
Our proposition can  be part of a good methodology for building future NLI datasets. 
The multi-faceted nature of the problems that exist in current NLI datasets  makes research that investigates these issues 
very important; 
the more the diagnostic tests we have, the more reliable the datasets will hopefully get. 
The fact that one type of testing (hypothesis only, word shuffling or word class dropping) 
does not eliminate all problems present in the datasets, highlights the need for a variety of diagnostic devices addressing different phenomena. 

We propose the following set of diagnostics as the minimum sanity check when developing new NLI datasets:
\begin{itemize}
    \item Hypothesis only baseline \citep{gururangan-etal-2018-annotation, poliak2018}
    \item Word-order shuffling \citep{Phametal:2020}
    \item Swapping premises and hypotheses  
    \citep{wang2019}
    \item Word class dropping (our proposed diagnostics)
\end{itemize}

\begin{figure}[t!]
\includegraphics[width=7.5cm]{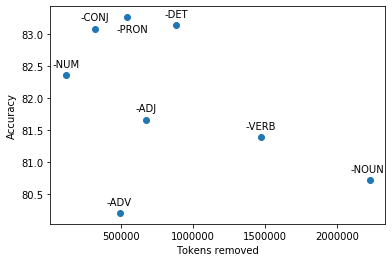}
  \caption{Comparison of BERT-{\tt base} model Accuracy vs Tokens removed. 
  The model is fine-tuned on the MNLI training data with instances 
  of a specific word class removed, 
  and evaluated on the original MNLI-matched development data.}
    \label{fig:accuracy_vs_tokens_removed}
\end{figure}

Returning to the specific findings of this paper, we performed an additional set of analysis aimed at identifying 
what the observed, relatively small, impact of the proposed modifications is due to. 
We explore whether the drop in performance can be explained by the (smaller or larger) number of tokens 
pertaining to the word class being removed. 
As can be seen in Figure \ref{fig:accuracy_vs_tokens_removed}, where we compare the accuracy of BERT and the number of tokens removed from the training data in each setting, this factor does not 
explain the obtained results. For example, there are only 492,895 occurrences of adverbs removed from the training set, but the delta to the original result is the highest (-3.53 points), whereas removing 886,966 determiners has only a small impact on accuracy (-0.59 points). This plot demonstrates the important role of content words in NLI prediction. 

\begin{figure}[t!]
\includegraphics[width=\columnwidth]{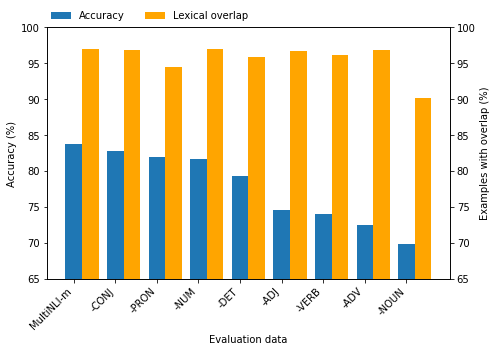}
  \caption{Comparison of model accuracy and lexical overlap 
  in the original MNLI test and the {\sc Corrupt-Test} sets. The models are fine-tuned on the original MNLI training data.} 
    \label{fig:lexical_overlap}
\end{figure}

 \citet{zhou-bansal-2020-towards} have shown that high lexical overlap between premises and hypotheses can guide models' predictions. We thus explore the extent to which our results can be explained by the amount of lexical overlap
in the {\sc Corrupt-Test} sets. We measure lexical overlap by counting the tokens shared by the premise and the hypothesis in a sentence pair.  The orange bars in the plot in Figure \ref{fig:lexical_overlap} 
show the amount of lexical overlap between premises and hypotheses (\%  
calculated over the total number of examples) in the original MNLI and the {\sc Corrupt-Test} test sets. The blue bars show the prediction accuracy obtained by BERT fine-tuned on the original MNLI data when evaluated on each test set. 
We observe that although there is a decrease in lexical overlap in some test sets (e.g., in {\sc mnli-noun}), there is no clear correlation between  lexical overlap and accuracy, which suggests that the model picks up on other cues that remain in the corrupted sentences for prediction. 

\section{Conclusion}

We propose a novel diagnostics suite for assessing the quality of datasets used for NLI model training and evaluation. We show that data corruption is an efficient way to estimate  
dataset quality and their potential to reflect the real language understanding capabilities of the models. 
Our results on the MNLI and ANLI datasets show that our methodology can successfully identify datasets of high or low quality, i.e. whether a dataset  triggers models' 
reasoning potential or rather allows them 
to rely on cues and other statistical biases for prediction. Our proposed tests can be used for assessing the quality of existing benchmarks used by the community and interpreting the results accordingly, and also to guide the development of new datasets addressing reasoning tasks. In this latter case, data corruption would  serve to identify whether a dataset construction methodology and the adopted annotation guidelines are on the correct track.

Lastly, although it would be interesting to compare a larger number of architectures, we leave this comparison for future work due to lack of space and also in order to not confuse the reader, given the large number of settings where experiments are  conducted. 
We also 
focus in this paper on 
the MNLI and ANLI datasets, since our main concern is to 
cover as many corruption settings 
as possible. Extending the current work to other models and NLU datasets is a natural next step 
for future research. We have made our code available to promote research in this direction.\footnote{\url{https://github.com/Helsinki-NLP/nli-data-sanity-check}} Additionally, since the present 
 work leaves open questions as regards the factors 
behind the high performance observed on the corrupted datasets, we plan to more thoroughly investigate the 
 cues and artefacts on which the models rely and which allow them 
 to perform well in these tasks.

\section*{Acknowledgments}
\vspace{1ex}
\noindent

\begin{wrapfigure}[]{l}{0pt}
\includegraphics[scale=0.3]{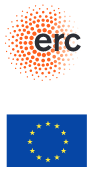}
\end{wrapfigure}

\noindent Marianna Apidianaki and Jörg Tiedemann are supported by the FoTran project, funded by the European Research Council (ERC) under the European Union’s Horizon 2020 research and innovation programme (grant agreement no.~771113). Stergios Chatzikyriakidis is supported by grant 2014-39 from the Swedish Research Council, which funds the Centre for Linguistic Theory and Studies in Probability (CLASP) in the Department of Philosophy, Linguistics, and Theory of Science at the University of Gothenburg. We thank the reviewers for their thoughtful comments and valuable suggestions. 

\bibliography{nli}
\bibliographystyle{acl_natbib}
 
\clearpage
\section*{Appendix}
\setcounter{figure}{0} \renewcommand{\thefigure}{A.\arabic{figure}}
\setcounter{table}{0} \renewcommand{\thefigure}{A.\arabic{table}}
Table \ref{table:example_correct} contains examples of sentence pairs from the {\sc mnli-noun} test set for which BERT predicted the correct labels.  
Table \ref{table:datasets} contains statistics for the number of tokens removed from the corrupted MNLI datasets. Table \ref{table:anli-stats} contains statistics for the number of tokens removed from the corrupted ANLI test sets. 
\begin{table*}[ht]
\small
\centering
        \begin{tabular}{p{1.7cm} p{6.5cm}  p{5.5cm}} 

\parbox{3.2cm}{ \bf Label} & \parbox{3.2cm}{\bf Premise} & \parbox{3.2cm}{\bf  Hypothesis} \\
\hline
\multirow{1}{*}contradiction & \em The intends that with appropriate in developing this. & \em The discourages to consult with any. \\
\hdashline
\multirow{1}{*}contradiction & \em Like and, warns, and Japanese are joined by yet locked in traditional. &\em and Japanese have no between them. \\
\hdashline
\multirow{1}{*}contradiction & \em To be sure, not all are. &\em  Every single is a. \\
\hline
\multirow{1}{*}entailment &\em  The, or Where the? & \em The of saving.\\
\hdashline
\multirow{1}{*}entailment &\em  In the original, is set up by his and then ambushed by a hostile named, and when he tries to answer with an eloquent ( is clenched. &\em  is out to get him.\\
\hdashline
\multirow{1}{*}entailment &\em  The other is retrospective and intended to help those who review to assess the of completed. &\em It is made to help the assess the of the.\\
\hline
\multirow{1}{*}neutral &\em  and uh it that takes so much away from your &\em  you away from your because it is more important to you.\\
\hdashline
\multirow{1}{*}neutral & \em The had been found in a in the. &\em  The that was in the was powdered. \\
\hdashline
\multirow{1}{*}neutral &\em In the other, the beat the. & \em The are a better.\\
            \hline
        \end{tabular}
    \caption{\label{table:example_correct} 
    Randomly selected sentence pairs from {\sc mnli-noun} test set for which BERT predicted the correct labels. 
    } 
\end{table*} 

\begin{table*}[ht!]
\small
\centering
        \begin{tabular}{l r r r | r r r} 
        
& \multicolumn{3}{c}{{\bf Training datasets}} & \multicolumn{3}{c}{{\bf Test datasets}} \\
\bf Configuration & \bf 
Premises & \bf 
Hypotheses &\bf  Total & \bf Premises & \bf Hypotheses & \bf  Total  \\ 
            \hline
{\sc mnli-num} & 119,587 & 44,289 & 163,876 & 3,100 & 1,133 & 4,233 \\
{\sc mnli-conj} 
& 320,210 & 76,466 & 396,676 & 7,584 & 1,874 & 9,458 \\
{\sc mnli-adv} 
& 492,895 & 237,250 & 730,145 & 11,777 & 5,862 & 17,639  \\
{\sc mnli-pron} & 543,968 & 301,293 & 845,261 & 13,060 & 7,466 & 20,526 \\
{\sc mnli-adj} & 677,095 & 302,652 & 979,747 & 16,162 & 7,562 & 23,724  \\
{\sc mnli-det} & 886,966 & 483,238 & 1,370,204 & 21,198 & 11,723 & 32,921 \\
{\sc mnli-verb} & 1,474,454 & 886,597 & 2,361,051 & 35,813 & 22,101 & 57,914  \\
{\sc mnli-noun} & 2,228,780 & 1,090,814 & 3,319,594 & 54,700 & 27,182 & 81,882 \\
{\sc mnli-noun-pron} & 2,772,748 & 1,392,107 & 4,164,855 & 67,760 & 34,648 & 102,408 \\
\hdashline
{\sc noun+pron+verb} & 4,501,189 & 2,166,146 & 6,667,335 & 109,325 & 53,647 & 162,972\\
{\sc noun+adv+verb} & 4,552,262 & 2,230,189 & 6,782,451 & 110,608 & 55,251 & 165,859\\
{\sc noun+verb} & 5,045,157 & 2,467,439 & 7,512,596 & 122,385 & 61,113 & 183,498\\
{\sc noun+verb+adj} & 4,368,062 & 2,164,787 & 6,532,849 & 106,223 & 53,551 & 159,774\\
{\sc noun+verb+adv+adj} & 3,875,167 & 1,927,537 & 5,802,704 & 94,446 & 47,689 & 142,135\\
            \hline
        \end{tabular}
    \caption{\label{table:datasets} Datasets formed by removing tokens from MNLI. The numbers correspond to number of tokens removed from the Premises and Hypotheses, and the total number of removed tokens.}
\end{table*}

\begin{table*}[ht]
\small
\centering
        \begin{tabular}{l r r r | r r r | r r r}
        & \multicolumn{3}{c}{{\bf R1}} & \multicolumn{3}{c}{{\bf R2}} & \multicolumn{3}{c}{{\bf R3}} \\
\bf Test dataset & \bf Premises & \bf Hypotheses & \bf Total& \bf Premises & \bf Hypotheses & \bf Total& \bf Premises & \bf Hypotheses & \bf Total \\
\hline
{\sc anli-noun} & 23,523 & 4,719 & 28,242 & 23,646 & 4,275 & 27,921 & 23,086 & 4,033 & 27,119 \\
{\sc anli-verb} & 6,057 & 1,657 & 7,714 & 6,155 & 1,668 & 7,823 & 11,281 & 2,258 & 13,539 \\
{\sc anli-pron} & 1,567 & 184 & 1,751 & 1,657 & 178 & 1,835 & 4,152 & 446 & 4,598 \\
{\sc anli-adj} & 2,827 & 514 & 3,341 & 2,783 & 495 & 3,278 & 3,525 & 625 & 4,150 \\
{\sc anli-adv} & 899 & 267 & 1,166 & 917 & 313 & 1,230 & 2,898 & 470 & 3,368 \\
{\sc anli-num} & 2,934 & 576 & 3,510 & 2,862 & 515 & 3,377 & 1,737 & 286 & 2,023 \\
{\sc anli-conj} & 1,816 & 161 & 1,977 & 1,897 & 122 & 2,019 & 2,073 & 142 & 2,215 \\
{\sc anli-det} & 5,631 & 1,195 & 6,826 & 5,669 & 1,086 & 6,755 & 7167 & 1,406 & 8,573 \\
\hline
        \end{tabular}
    \caption{\label{table:anli-stats}Datasets formed by removing tokens from ANLI test sets. The numbers correspond to number of tokens removed from the Premises and Hypotheses, and the total number of removed tokens for the three datasets (rounds).}
\end{table*}

\end{document}